\documentclass[conference]{IEEEtran}
\usepackage{cite}
\usepackage{amsmath,amssymb,amsfonts}
\usepackage{algorithmic}
\usepackage{graphicx}
\usepackage{textcomp}
\usepackage{microtype}
\usepackage{xcolor}
\usepackage{array} 
\usepackage{booktabs} 
\def\BibTeX{{\rm B\kern-.05em{\sc i\kern-.025em b}\kern-.08em
    T\kern-.1667em\lower.7ex\hbox{E}\kern-.125emX}}
\begin{document}

\title{Emotional Intelligence Through Artificial Intelligence : NLP and Deep Learning in the Analysis of Healthcare Texts\\
  \thanks{}
}

\author{\IEEEauthorblockN{Prashant Kumar Nag}
  \IEEEauthorblockA{\textit{Department of Computer Applications} \\
    \textit{Maulana Azad National Institute of Technology (MANIT)}\\
    Bhopal, India \\
    pn.193120001@manit.ac.in}
  \and
  \IEEEauthorblockN{Amit Bhagat}
  \IEEEauthorblockA{\textit{Department of Computer Applications} \\
    \textit{Maulana Azad National Institute of Technology (MANIT)}\\
    Bhopal, India \\
    am.bhagat@gmail.com}
  \and
  \IEEEauthorblockN{R. Vishnu Priya}
  \IEEEauthorblockA{\textit{Department of Computer Applications} \\
    \textit{National Institute of Technology, Tiruchirappalli}\\
    Tiruchirappalli, India \\
    vishnupriya@nitt.edu}
  \and
  \IEEEauthorblockN{Deepak Kumar Khare}
  \IEEEauthorblockA{\textit{Department of Computer Applications} \\
    \textit{Maulana Azad National Institute of Technology (MANIT)}\\
    Bhopal, India\\
    deepakkhare.kd@gmail.com}
}

\maketitle

\begin{abstract}
  This manuscript presents a methodical examination of the utilization of
  Artificial Intelligence (AI) in the assessment of emotions in texts
  related to healthcare, with a particular focus on the incorporation of
  Natural Language Processing (NLP) and deep learning technologies. We
  scrutinize numerous research studies that employ AI to augment sentiment
  analysis, categorize emotions, and forecast patient outcomes based on
  textual information derived from clinical narratives, patient feedback
  on medications, and online health discussions. The review demonstrates
  noteworthy progress in the precision of algorithms used for sentiment
  classification, the prognostic capabilities of AI models for
  neurodegenerative diseases, and the creation of AI-powered systems that
  offer support in clinical decision-making. Remarkably, the utilization
  of AI applications has exhibited an enhancement in personalized therapy
  plans by integrating patient sentiment and contributing to the early
  identification of mental health disorders. There persist challenges,
  which encompass ensuring the ethical application of AI, safeguarding
  patient confidentiality, and addressing potential biases in algorithmic
  procedures. Nevertheless, the potential of AI to revolutionize
  healthcare practices is unmistakable, offering a future where healthcare
  is not only more knowledgeable and efficient but also more empathetic
  and centered around the needs of patients. This investigation
  underscores the transformative influence of AI on healthcare, delivering
  a comprehensive comprehension of its role in examining emotional content
  in healthcare texts and highlighting the trajectory towards a more
  compassionate approach to patient care. The findings advocate for a
  harmonious synergy between AI's analytical capabilities and the human
  aspects of healthcare, guaranteeing that technological advancements are
  aligned with the emotional well-being of patients.
\end{abstract}


\section{Introduction}

The integration of Artificial Intelligence (AI) into healthcare represents a transformative leap forward in enhancing patient care through the nuanced analysis of clinical narratives. Within the realm of emotional intelligence, AI technologies, notably Natural Language Processing (NLP) and Deep Learning, have begun to exert an indispensable influence. This paper, "Emotional Intelligence Through Artificial Intelligence: NLP and Deep Learning in the Analysis of Healthcare Texts," systematically reviews the significant strides and applications AI has made in the emotional analysis of health-related texts, underlining the pivotal contributions each referenced study brings to the field.

One of the seminal contributions to this domain is the incorporation of lexicon-based sentiment classification methods with logistic regression, which has revolutionized sentiment analysis in clinical narratives, aiding clinicians in interpreting complex emotional data derived from patient records ~\cite{sanglerdsinlapachai2021ImprovingSentimentAnalysis}. Building upon this foundation, the development of cutting-edge models like the Hybrid Value-Aware Transformer marks a milestone, enhancing the predictive capabilities for conditions such as Alzheimer’s disease ~\cite{shao2023HybridValueAwareTransformer} and showcasing the real-world impact of AI in predicting and managing complex health conditions.

Moreover, the innovative application of sentiment analysis on drug review data serves as a paradigm of personalized therapy, where patient feedback is intricately woven into the clinical decision-making fabric ~\cite{hiremath2022EnhancingOptimizedPersonalized}. This work stands as a testament to AI's capacity for tailoring patient-centric treatment strategies, demonstrating the direct influence AI has on enhancing therapeutic outcomes.

The scope of AI's influence extends beyond the direct analysis of patient data, reaching into psychometric analyses and online patient review evaluations ~\cite{zhu2021OnlineCriticalReview, vij2018AutomatedPsychometricAnalyzer}. These contributions are instrumental in refining patient care, managing satisfaction, and providing new avenues for addressing mental health disorders. The research not only illustrates AI's prowess in gleaning insights from an array of data sources but also underscores its role in elevating healthcare delivery to new heights of efficiency and personalization.

Additionally, cognitive network science emerges as a powerful tool in mental healthcare through the work of ~\cite{joseph2023CognitiveNetworkNeighborhoods}, providing groundbreaking methodologies to decipher the emotional content within textual data. This work contributes significantly to therapeutic interventions by offering novel insights into the emotional well-being of individuals.

Furthermore, the utilization of machine learning and NLP to address critical public health issues such as intimate partner violence and depression underscores the societal value of AI ~\cite{al-garadi2022NaturalLanguageModel, yang2020BigDataAnalytics, khan2023HybridMachineLearning}. These studies not only enhance public health surveillance and intervention strategies but also mark a leap forward in our collective approach to mental wellness.

In addressing the crucial role of AI in healthcare, this paper not only presents a comprehensive review of current methodologies and applications but also highlights the transformative contributions of these technologies in fostering a more informed, empathetic, and effective healthcare ecosystem.

\section{Methodology}

\subsection{Search Terms and Databases}

Several research subjects were found within the following domains: study
of emotions, artificial intelligence, health, and text connected to
health. The search method, inclusion criteria, and selection of relevant
publications were performed in accordance with the Preferred Reporting
Items for Systematic Reviews and Meta-Analyses (PRISMA) guidelines as
outlined by ~\cite{peng2020HumanMachineDialogue}. The systematic search
was conducted in the Pubmed/Medline, Web of Science, and Scopus
databases.The search terms for different dtabases are mentioned in the
Table~\ref{tbl:database}.

\begin{table}[!h]
  \caption{Search Queries for each Databases }\tabularnewline
  \label{tbl:database}
  \centering
  \begin{tabular}{l>{\raggedright\arraybackslash}p{5.5cm}}
    \toprule
    Databases     & Search Query                                                                                                                                                                                               \\
    \midrule
    ScienceDirect & ( "Artificial Intelligence" AND "Deep Learning" ) AND ( "Emotion" AND "Text Analysis" AND "Natural Language Processing" ) AND ( "Healthcare" AND "Medical Texts" OR "Clinical Data" OR "Patient Records" ) \\
    Scopus        & ( "Artificial Intelligence" OR "Deep Learning" ) AND ( "Emotion" AND "Text Analysis" OR "Natural Language Processing" ) AND ( "Healthcare" AND "Medical Texts" OR "Clinical Data" OR "Patient Records" )   \\
    PubMed        & ("Artificial Intelligence" OR "Deep Learning") AND ("Emotion" OR "Text Analysis" OR "Natural Language Processing") AND ("Healthcare" AND "Medical Texts" OR "Clinical Data" OR "Patient Records")          \\
    \bottomrule
  \end{tabular}
\end{table}

\subsection{Criteria for including studies}

In order to identify relevant publications for our investigation, we
utilized the following criteria:

\begin{itemize}
  \item
        This study focuses on articles authored in the English language and
        published in peer-reviewed journals throughout the period from January
        1, 2013, to October 31, 2023.
  \item
        The primary emphasis of the articles should center around the
        utilization of Artificial Intelligence in the realm of emotional
        analysis.
  \item
        The selected papers must incorporate health-related materials as the
        major source of data.
  \item
        The user exhibits a preference for scholarly papers that analyze brief
        texts or materials sourced from Clinical narratives, drug reviews,
        Psychological-patient Texts and Medical Text datasets.
\end{itemize}

These criteria guarantee that the selected articles are pertinent,
up-to-date, and primarily focus on the convergence of artificial
intelligence (AI) and emotional analysis within the healthcare domain,
particularly emphasizing the analysis of brief or social media-derived
texts.

\subsection{Data collection and Management}

This systematic review used a comprehensive data gathering and
management technique to include relevant and high-quality research. The
data collection procedure has numerous steps:

\sloppy
\begin{itemize}
  \item
        Search and Select Databases: PubMed, Scopus, and Web of Scienece
        academic databases were meticulously searched. Studies published
        between January 1, 2013, and October 31, 2023 were filtered using
        ``Artificial Intelligence``, ``Emotional Analysis``, ``Health-related
        Texts``, and ``Social Media`` keywords.
  \item
        Inclusion/Exclusion Criteria: English, peer-reviewed, AI-based
        emotional analysis, and health-related or short social media messages
        were the selection criteria.
  \item
        Extracting Data: Each selected article provided publication year,
        methodology, AI approaches, type of health-related text studied,
        conclusions, and limitations. Tabulated data made comparison and
        analysis easier.
  \item
        Assessing Quality: Each work was assessed for methodological rigor and
        research relevance. This review ensured data reliability and validity.
  \item
        Manage Data: Reference management software \textbf{Zotero} organised
        all data. This made knowledge retrieval, review, and synthesis easier
        during research.
\end{itemize}
\fussy

\section{Results and Discussion}

\begin{figure}[!h]
  \centering
  \includegraphics[width=3.4in]{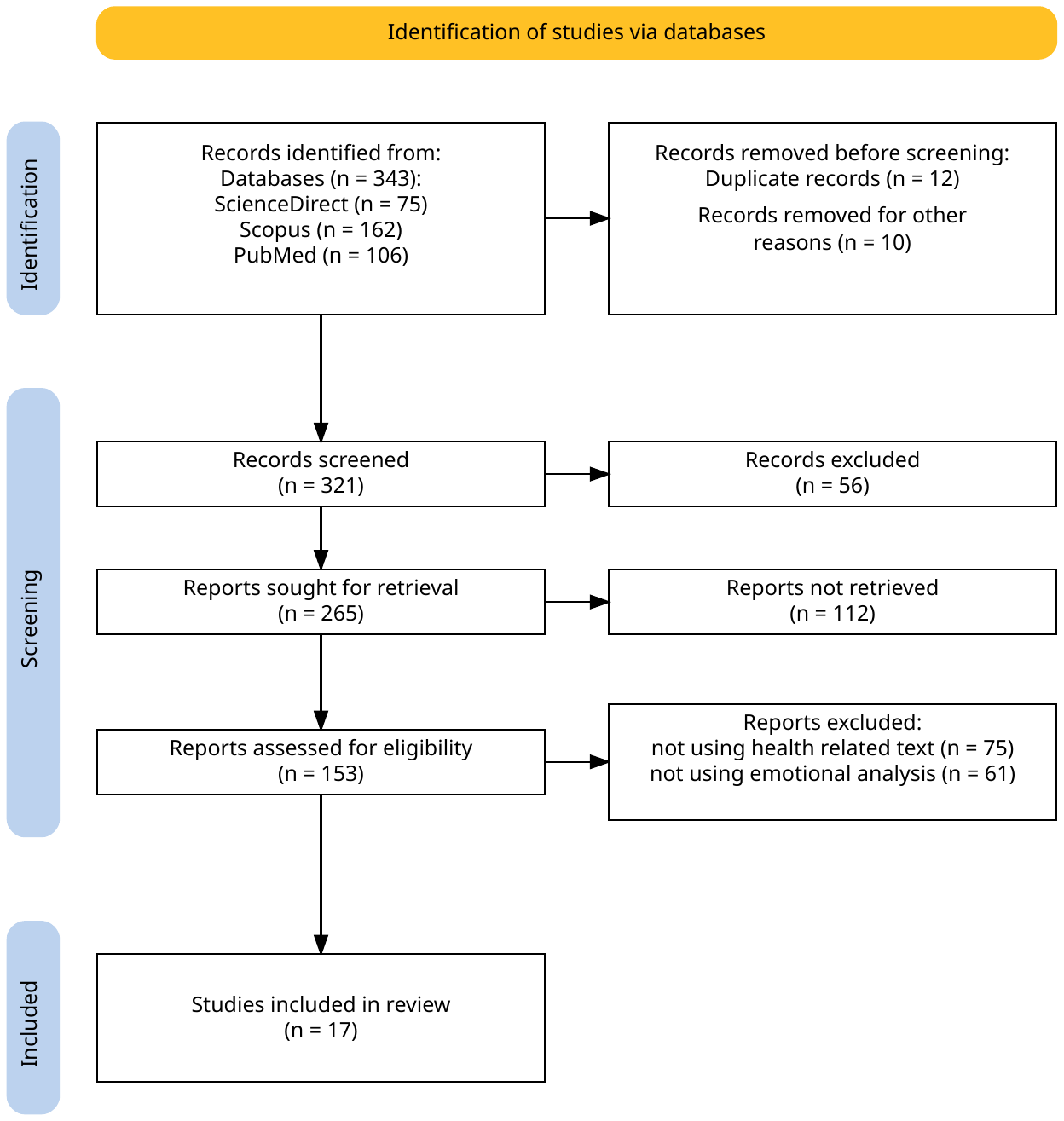}
  \caption{The PRISMA flowchart}
  \label{fig:prisma}
\end{figure}

The preliminary search yielded 343 records, depicted in the PRISMA flow
diagram (refer to Figure ~\ref{fig:prisma}). From this total, 12 records were
identified as duplicates and subsequently eliminated. An additional 10
records were discarded for other reasons. This process resulted in the
screening of 321 records. Following this, 56 records were excluded,
leading to 265 records being selected for further retrieval. However,
112 of these records could not be retrieved. Ultimately, the search
process identified 153 articles as potentially relevant to the study.
After a thorough review, 17 full-text articles were determined to meet
all the specified eligibility criteria and were included in the final
analysis. PRISMA Flow Diagram and study selection.

Recent advancements in AI have significantly bolstered emotional analysis in healthcare,
as illustrated by the studies in this review.
The Author ~\cite{sanglerdsinlapachai2021ImprovingSentimentAnalysis} has enhanced sentiment classification in clinical narratives,
allowing healthcare professionals to make more informed treatment decisions.
Similarly, the HVAT model by ~\cite{shao2023HybridValueAwareTransformer} excels in predicting Alzheimer's disease,
hinting at its broader clinical applicability.

~\cite{hiremath2022EnhancingOptimizedPersonalized} has developed a clinical decision support system utilizing
drug review data for personalized therapy, indicating AI's promise in integrating patient sentiments into clinical decision-making.
~\cite{zhu2021OnlineCriticalReview}'s work transferring techniques from the hospitality to healthcare industry potentially improves patient care through the analysis of online reviews.

The untapped potential in healthcare for sentiment analysis and emotional recognition is highlighted by \cite{vij2018AutomatedPsychometricAnalyzer}, indicating a significant impact on the treatment of mental health disorders. The study by \cite{joseph2023CognitiveNetworkNeighborhoods} on cognitive network science provides mental health professionals with novel tools to understand emotional states, potentially aiding in suicide prevention and mental health interventions.

In the realm of public health surveillance, \cite{al-garadi2022NaturalLanguageModel}'s machine learning and NLP model for IPV detection on social media can enhance healthcare surveillance and aid IPV victims. \cite{yang2020BigDataAnalytics}'s framework for depression detection from social media behavior offers new prospects for early detection and intervention.

Furthermore, the hybrid machine-learning models by \cite{khan2023HybridMachineLearning} demonstrate high accuracy in sentiment analysis for depression detection, suggesting their use in large-scale mental health monitoring in healthcare systems. The multi-modal framework by \cite{kaur2023SentimentAnalysisLinguistic}, which combines image classification with sentiment analysis, shows significant improvement in medical image classification and presents a more comprehensive understanding of complex cases.

The DC-LSTM model by \cite{liang2021ImprovedDoubleChannel} for classifying complex medical texts suggests promising applications for accurate patient triage and treatment. The study by \cite{jimenez-zafra2019HowWeTalk} provides insights into the complex sentiment classification in medical forums, emphasizing the unique linguistic features in drug reviews.

The lexicon developed by \cite{mondal2018RelationExtractionMedical} for medical concept categorization, incorporating sentiment analysis, aids in developing medical ontologies and recommendation systems. The feature evaluation for polarity classification by \cite{carrillo-de-albornoz2018FeatureEngineeringSentiment} enhances understanding of patient-authored content in e-health forums.

The classification approach for chronically ill patients' health records by \cite{bromuri2014MultilabelClassificationChronically} facilitates clinical decision-making. \cite{straw2020ArtificialIntelligenceMental}'s work on biases in NLP models highlights the importance of mitigating biases in digital mental healthcare.

Lastly, the global health perspective by \cite{hadley2020ArtificialIntelligenceGlobal} underscores AI's role in improving medical AI in lower-income countries, advocating for an effective framework for AI technologies in global health initiatives.

These studies collectively underscore the transformative impact of AI in healthcare, from enhancing diagnostic precision to offering personalized patient care. They reflect a future in which AI not only streamlines clinical practices but also addresses the emotional and psychological aspects of patient health, thereby revolutionizing healthcare delivery.

NLP and Deep Learning Applications in Emotional Analysis of Healthcare
Texts is shows in Table~\ref{tbl:qimprovement}

\begin{table*}
  \centering
  \caption{NLP and Deep Learning Applications in Emotional Analysis of Healthcare Texts}
  \label{tbl:qimprovement}
  \begin{tabular}{>{\raggedright\arraybackslash}p{.4cm} >{\raggedright\arraybackslash}p{2.25cm} >{\raggedright\arraybackslash}p{2.5cm} >{\raggedright\arraybackslash}p{3.5cm} >{\raggedright\arraybackslash}p{3cm} >{\raggedright\arraybackslash}p{3.5cm}}
    \toprule
    Study References                                           & AI Technology Used                                                              & Focus Area                                                                  & Data Source                                                                                                   & Key Contributions                                                                              & Relevance to Healthcare                                                                                                                  \\
    \midrule
    \cite{sanglerdsinlapachai2021ImprovingSentimentAnalysis}   & Machine Learning, Natural Language Processing                                   & Sentiment Analysis, Clinical Narrative Analysis                             & Clinical narratives from patient records                                                                      & Enhanced sentiment classification to 0.882 accuracy with SentiWordNet and UMLS semantics.      & Enhances sentiment analysis in clinical narratives, aiding healthcare professionals in patient care decisions.                           \\
    \cite{shao2023HybridValueAwareTransformer}                 & Hybrid Value-Aware Transformer (HVAT), a Transformer-based deep neural network. & Prediction of Alzheimer’s disease and related dementias.                    & A case-control dataset containing longitudinal and non-longitudinal clinical data.                            & Improved Alzheimer’s prediction using a novel DNN with "clinical tokens."                      & HVAT model's potential in processing clinical data suggests wider applications in disease prediction.                                    \\
    \cite{hiremath2022EnhancingOptimizedPersonalized}          & Machine learning methods like SVM, Random Forest Classification.                & Sentiment analysis on drug review                                           & Drug review data used to determine the sentiments of patients towards medicines, treatments, etc.             & Superior patient opinion classification on medications using SVM for decision-making.          & Improves clinical decisions by integrating patient feedback on medication into treatment planning.                                       \\
    \cite{zhu2021OnlineCriticalReview}                         & Deep Learning-based Natural Language Processing (NLP).                          & Response strategies of firms within the hospitality industry                & An extensive set of 110,146 online reviews from a hospitality chain's operation collected over seven years.   & Identified key predictors for online review criticality, boosting algorithmic accuracy.        & Transferable analysis methods from hospitality to healthcare could lead to improved patient care.                                        \\
    \cite{vij2018AutomatedPsychometricAnalyzer}                & Deep Learning and Natural Langauge Processing.                                  & Psychological issues like depression, bipolar disorders, and schizophrenia. & The data sources mentioned include psychologist-patient written texts and conversations.                      & Utilized sentiment analysis to improve medical record visualization for healthcare.            & Untapped sentiment analysis and emotional recognition in healthcare could benefit mental health care.                                    \\
    \cite{joseph2023CognitiveNetworkNeighborhoods}             & Natural Langugage Processing and Machine Learning.                              & Emotions in suicide notes and mental health conversations                   & The study transformed 142 suicide notes and 77,000 Reddit posts.                                              & Quantified emotions across digital platforms, uncovering diverse emotional patterns.           & Offers a network tool for mental health care, enabling professionals to assess patients' emotional states digitally.                     \\
    \cite{al-garadi2022NaturalLanguageModel}                   & Machine Learning(ML) and NLP.                               & Public health surveillance.                                                 & A total of 6,348 tweets were manually reviewed and annotated.                                                 & Set IPV tweet annotation standards and developed a high-accuracy NLP model.                    & Supports IPV victim intervention and healthcare surveillance, providing insights for resource allocation.                                \\
    \cite{yang2020BigDataAnalytics}                            & Machine Learning and NLP.                               & Identification of depression in social media users.                         & Dataset obtained from Facebook.                                                                               & Integrated pragmatic and social data in a new framework for depression detection.              & Provides early depression detection tools, aiding in diagnosis and mental health outcome improvement.                                    \\
    \cite{khan2023HybridMachineLearning}                       & Machine Learning and NLP.                               & Identifying signs of depression.                                            & Publicly available datasets of sentiment tweets.                                                              & Demonstrated high accuracy in depression detection.                                            & Linguistic analysis inclusion enhances diagnostic accuracy in medical image classification.                                              \\
    \cite{kaur2023SentimentAnalysisLinguistic}                 & Deep learning and Bag of Words (BoW).     & Classification of medical images.                                           & Gastral images and related text obtained from a gastroenterologist.                                           & Advanced image classification with LSTM and BoW, integrating sentiment analysis.               & DC-LSTM model aids in accurate triage and treatment with its complex text classification capabilities.                                   \\
    \cite{liang2021ImprovedDoubleChannel}                      & Deep Learning, LSTM.                                                            & Classification of complex medical texts, Chinese medical diagnosis.         & cMedQA (a Chinese medical diagnosis dataset) and Sentiment140 (a sentiment analysis dataset)                  & Double Channel LSTM with hybrid attention outshines standard CNN-LSTM models.                  & Emphasizes the need for specific language consideration in medical sentiment analysis for patient satisfaction insights.                 \\
    \cite{jimenez-zafra2019HowWeTalk}                          & Machine learning                                                                & Sentiment analysis in medical forums                                        & Two Spanish corpora: (i) DOS (drug reviews) and (ii) COPOS (patients’ opinions about physicians).             & Showed complexity in sentiment classification of drug reviews with mixed language.             & Automated system enhances clinical decision-making with insights into medical term semantics and supports ontology development.          \\
    \cite{mondal2018RelationExtractionMedical}                 & Machine Learning and Natural Language Processing(NLP).                               & Automated extraction of semantic relations.                                 & Text corpora in healthcare services (specific datasets are not mentioned in the provided abstract).           & Created a lexicon merging sentiment analysis for medical concept categorization.               & Essential for accessing accurate e-health information and analyzing patient-authored medical content.                                    \\
    \cite{carrillo-de-albornoz2018FeatureEngineeringSentiment} & Machine Learning and Natural Language Processing(NLP).                               & Sentiment analysis in e-health forums.                                      & Annotated dataset of over 3500 posts from online health forums related to breast cancer, Crohn's disease .etc & Polarity classification in patient texts improved by word embeddings and polar facts.          & Confirms BoW's effectiveness in medical record classification and the utility of dimensionality reduction for clinical decision support. \\
    \cite{bromuri2014MultilabelClassificationChronically}      & Bag of Words (BoW), Machine Learning(ML)                                            & Multi-label classification of illnesses in chronically ill patients.        & Portavita dataset with 525 diabetes type 2 patients, and the MIMIC II dataset with 2,635 patients data        & Introduced a BoW and dimensionality reduction method for time series medical data.             & Important for healthcare professionals to address AI-induced health inequalities and decision-making biases.                             \\
    \cite{straw2020ArtificialIntelligenceMental}               & Deep Learning and Natural Langauge Processing.                                  & Mental health and language-based models.                                    & data extracted from text corpora such as social media posts, online forums, blogs, and chat rooms.            & Identified biases in 'GloVe' and 'Word2Vec' for mental health, signaling research needs.       & Offers a strategic AI framework for global health, promoting medical AI improvements in low-income regions.                              \\
    \cite{hadley2020ArtificialIntelligenceGlobal}              & Machine Learning and Natural Langauge Processing (NLP).                               & Improve global health outcomes                                              & AI-based Global Health Initiatives (GHIs)                                                                     & Proposed a sustainable AI framework aimed at enhancing global health research. & Critically assesses AI’s impact on global health, advocating a framework to ensure equitable AI benefits in healthcare.                  \\
    \bottomrule
  \end{tabular}
\end{table*}

\section{Limitations}

The exploration of AI's role in emotional analysis within health-related
texts has advanced significantly, yet it is not without its limitations.
For instance:

Data Source Bias: The reliance on social media platforms like Twitter
and \cite{benrouba2023EmotionalSentimentAnalysis} Facebook for
emotional analysis could introduce bias, as these platforms may not
accurately reflect the broader population's emotional states.

Model Complexity: While the addition of irony, sarcasm, and emoji
detection \cite{brezulianu2022NotOurFeeling} enriches the analysis, the
increased complexity of sentiment analysis models can also obscure
interpretability and hinder the practical application in clinical
settings.

Aspect Granularity: Efforts to improve aspect-level granularity
\cite{gan2019AdaptiveLearningEmotion} are promising, yet current models
may still overlook nuanced emotional expressions that do not fit
predefined categories.

Cultural and Contextual Sensitivity: The application of algorithms like
VADER and LDA may not fully account for cultural and contextual
variations in language use, potentially leading to misinterpretation of
sentiments.

Generalizability: Studies utilizing specific datasets, such as user
conversation data
\cite{inkster2018EmpathydrivenConversationalArtificial} , may not
generalize to other forms or sources of health-related texts.

Deep Learning Limitations: The use of basic categorization models,
including deep learning \cite{liu2022MultistageDeepTransfer} , requires
extensive data and computational resources, which may not be feasible
for all research or clinical entities.

Chatbot Responsiveness: Although improvements in chatbot interactions
\cite{rahmanti2022SlimMeChatbotArtificial} aim to increase accuracy,
these systems can still struggle with the subtleties of human
conversation and may not always provide appropriate or empathetic
responses.

Neural Network Challenges: Employing multi-layer artificial neural
networks \cite{rechowicz2023UseArtificialIntelligence} offers enhanced
classification, but the black-box nature of these models can make it
difficult to trace how decisions are made, complicating clinical
accountability.

User Interface Adaptability: Improvements in app navigation and
interface \cite{shan2022PublicTrustArtificial} must continually adapt
to the evolving user expectations and accessibility requirements, which
is a continual developmental challenge.

Performance Metrics: Lastly, reliance on traditional performance metrics
such as accuracy, precision, and recall
\cite{yadollahi2017CurrentStateText} may not always capture the quality
of emotional analysis from a human-centric perspective.

\section{Conclusion}

The exploration of AI's capabilities in the realm of emotional analysis
within healthcare texts has provided valuable insights into the evolving
interface of technology and patient care. The systematic review
presented in this paper underscores the significant strides made in
sentiment analysis, predictive modeling, and decision support systems
through the integration of advanced AI technologies like NLP and deep
learning.

The convergence of AI with emotional intelligence heralds a new era in
healthcare - one where nuanced patient sentiments can be accurately
interpreted, leading to more empathetic and personalized care. Studies
such as those by
\cite{sanglerdsinlapachai2021ImprovingSentimentAnalysis} and
\cite{shao2023HybridValueAwareTransformer, r.2023TextbasedEmotionRecognition}
have demonstrated the practical efficacy of AI in enhancing sentiment
classification and predicting disease, respectively. These advancements
not only augment clinical decision-making but also broaden the scope of
preemptive healthcare practices.

Moreover, the adaptability of AI methods across different industries, as
illustrated by
\cite{zhu2021OnlineCriticalReview, nag2021ContextualBIDirectionalAttention}
, along with the potential of sentiment analysis and emotional
recognition in healthcare settings
\cite{vij2018AutomatedPsychometricAnalyzer}, indicate that AI's
application is both versatile and impactful. The use of cognitive
network science to understand emotional expressions
\cite{joseph2023CognitiveNetworkNeighborhoods} and the employment of
social media analytics for public health surveillance
\cite{al-garadi2022NaturalLanguageModel} further exemplify the breadth
of AI's applicability.

In conclusion, this paper reveals that while AI's journey in healthcare
is progressive, it is accompanied by challenges that require careful
navigation. Ethical considerations, data privacy, and the need to
address algorithmic biases are critical aspects that must be diligently
managed. As we move forward, the focus should be on leveraging AI to
complement the human aspects of healthcare ensuring that technology acts
as an ally to both healthcare professionals and patients. The future of
healthcare lies in a balanced synergy of artificial and emotional
intelligence, where AI empowers care providers to meet not just the
clinical but also the emotional needs of their patients.

\bibliography{bibliography}
\end{document}